\documentclass[letterpaper, 10 pt, conference]{ieeeconf}  % Comment this line out if you need a4paper

\usepackage[T1]{fontenc}
\usepackage[utf8]{inputenc}
\usepackage{authblk}
\usepackage{graphicx}
\usepackage[spaces,hyphens]{url}
\usepackage[linesnumbered,ruled]{algorithm2e}
\usepackage{array}
\usepackage{mathtools}
\usepackage{subcaption}
\usepackage{amssymb}
\usepackage{amsmath}
\usepackage{diagbox}
\usepackage{fourier} 
\usepackage{makecell}
\usepackage{hyperref}
\usepackage{stackengine}
\usepackage{multirow}
\usepackage{sidecap}
\usepackage[flushleft]{threeparttable} % http://ctan.org/pkg/threeparttable
\usepackage{booktabs,caption}
\usepackage{xcolor, soul}
\sethlcolor{yellow}

\IEEEoverridecommandlockouts                              % This command is only needed if 
\title{\LARGE \bf
Next-Best-View Prediction for Active Stereo Cameras and Highly Reflective Objects
}

\author{Jun Yang and Steven L. Waslander
\thanks{This work was supported by Epson Canada Ltd.}
\thanks{Authors are with University of Toronto Institute for Aerospace Studies and Robotics Institute.
        {\tt\footnotesize junyang.yang@mail.utoronto.ca, steven.waslander@robotics.utias.utoronto.ca}}
}

\begin{document}

\maketitle
\thispagestyle{empty}
\pagestyle{empty}

%%%%%%%%%%%%%%%%%%%%%%%%%%%%%%%%%%%%%%%%%%%%%%%%%%%%%%%%%%%%%%%%%%%%%%%%%%%%%%%%
\begin{abstract}
Depth acquisition with the active stereo camera is a challenging task for highly reflective objects. When setup permits, multi-view fusion can provide increased levels of depth completion. However, due to the slow acquisition speed of high-end active stereo cameras, collecting a large number of viewpoints for a single scene is generally not practical. In this work, we propose a next-best-view framework to strategically select camera viewpoints for completing depth data on reflective objects. In particular, we explicitly model the specular reflection of reflective surfaces based on the Phong reflection model and a photometric response function. Given the object CAD model and grayscale image, we employ an RGB-based pose estimator to obtain current pose predictions from the existing data, which is used to form predicted surface normal and depth hypotheses, and allows us to then assess the information gain from a subsequent frame for any candidate viewpoint. Using this formulation, we implement an active perception pipeline which is evaluated on a challenging real-world dataset. The evaluation results demonstrate that our active depth acquisition method outperforms two strong baselines for both depth completion and object pose estimation performance.
\end{abstract}

%%%%%%%%%%%%%%%%%%%%%%%%%%%%%%%%%%%%%%%%%%%%%%%%%%%%%%%%%%%%%%%%%%%%%%%%%%%%%%%%
\section{INTRODUCTION}
Reliable depth data acquisition is an important problem in many robotic applications. For example, in robotic grasping \cite{yan2020fast}, accurate 6D object pose estimation is generally required prior to the grasping execution, and its performance relies heavily on the input depth data quality. To acquire reliable depth images, the active stereo (AS)-based camera is widely used due to its high accuracy and resolution \cite{jang2013structured}. The AS-based camera employs a light projector to simplify the stereo matching problem, and can be further divided into the conventional active stereo camera (ASC) and the structured light camera (SLC). The ASC is equipped with two cameras, it first projects random patterns onto objects to provide visual features, and finds camera-camera correspondences. In comparison, SLC requires well-designed patterns (e.g., sinusoidal fringe) for encoding spatial information, and stereo matching is performed for projector-camera correspondences.

\begin{figure}[t]
\centering
\begin{subfigure}{0.48\textwidth}
  \includegraphics[width=\linewidth]{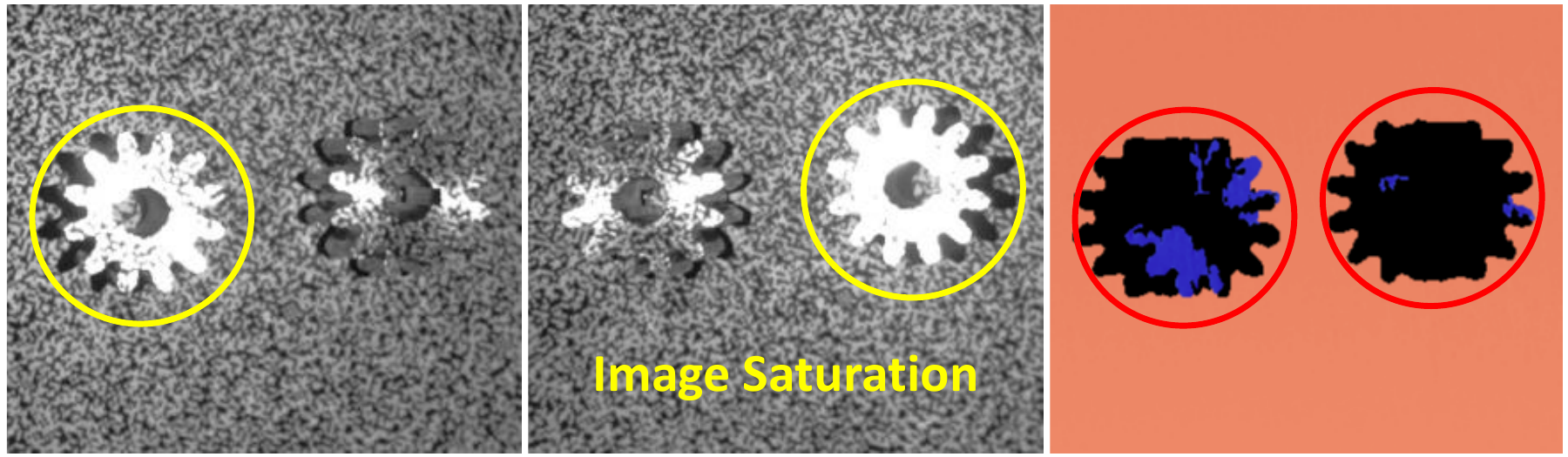}
  \vspace{-1.05\baselineskip}
  \label{fig1a}
\end{subfigure}
\begin{subfigure}{0.48\textwidth}
    \includegraphics[width=\linewidth]{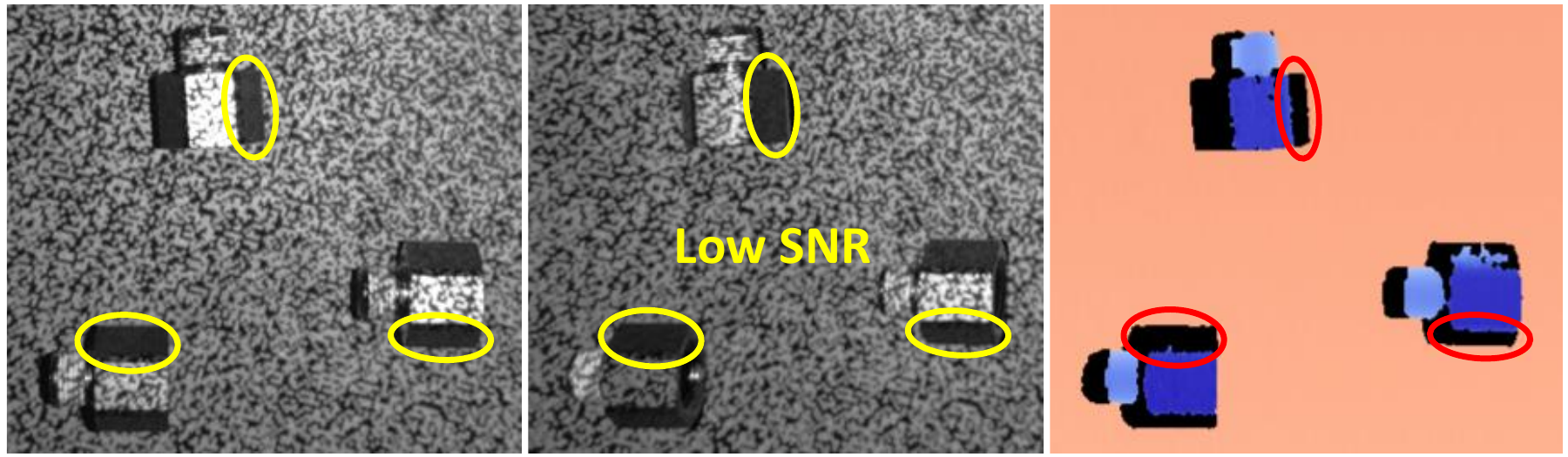}
    \vspace{-0.9\baselineskip}
    \label{fig1b}
\end{subfigure}
\caption{Missing depth data for reflective object surfaces using a conventional active stereo camera (ASC). Top: missing depth data caused by image saturation. Bottom: missing depth data caused by low SNR. From left to right: pattern projected stereo pair and depth map.}
\label{fig1}
\end{figure}

The AS-based camera excels when imaging objects with diffuse surfaces that have roughly Lambertian reflection. However, one challenge of the AS-based camera is the missing depth measurement for reflective objects, which are common in the real world (e.g., shiny metal parts). This situation with highly reflective surfaces is illustrated in Figure \ref{fig1}. Due to specular reflection, a high proportion of the incident illumination is reflected, either directly back to the camera resulting in image saturation, or in a separate direction, completely missing the camera and resulting in a low signal-to-noise ratio (SNR).  Both effects can result in a failure to measure the depth to the surface with the AS-based camera.

To overcome this problem, the high dynamic range (HDR) technique was developed \cite{zhang2009high, ekstrand2011autoexposure, yu2019adaptive, lin2017review, feng2018high}. The HDR approach involves measuring the object with multiple exposures to avoid image saturation. However, its acquisition is usually time-consuming and cannot address the low SNR issue. An alternative solution is to leverage the power of the neural network to directly predict the depth value at the missing areas \cite{zhang2018deep, chai2020deep, sajjan2020clear, luo2021self}. However, the accuracy of these approaches is generally low.

When the application requirements permit, we can place the camera at different viewpoints and fuse multi-view acquired depth maps \cite{yang2021probabilistic, wang2019highly}. However, due to the slow acquisition speed of the AS-based camera ($\textasciitilde 1\;$fps), capturing a large number of viewpoints is not practical. In this paper, we propose an active vision approach, and predict the next-best-view (NBV) to complete the reference view's depth data for reflective surfaces. Our proposed method is composed of two main parts: a) surface reflectance modeling, b) NBV prediction for depth completion. In the first part, we use the Phong reflection model \cite{phong1975illumination} and a photometric response function \cite{debevec2008recovering} to predict pattern image intensity for reflective surfaces. The predicted pixel intensity is then used to estimate the depth-sensing probability. In the second part, we use an RGB-based object pose estimator to obtain a predictive model of the environment, including surface normal and depth hypotheses at unobserved depth pixel locations. We then integrate the estimates with our reflection model to assess the information gain for each candidate viewpoint. We evaluate our framework on the challenging ROBI dataset \cite{yang2021robi}, showing that our NBV framework can predict information gain from viewpoints accurately, and achieves robust depth completion performance relative to two strong baselines. In summary, our work makes the following contributions:

\begin{itemize}
    \item A surface reflection model to estimate the depth-sensing probability on reflective object surfaces. The reflection model comprises of a) a photometric response function to recover sensor radiance, b) the calibration of Phong model parameters.
    \item A method to estimate prior information of the objects, such as surface normal and depth hypotheses.
    \item An active vision system that integrates the reflection model and scene priors to predict the NBV for the reflective object's depth completion.
\end{itemize}

The rest of the paper is structured as follows. Section \ref{sec2} reviews the literature. Section \ref{sec3} describes our surface reflection model. Section \ref{sec4} explains the NBV planner. Section \ref{sec5} presents the evaluation results, and section \ref{sec6} concludes the paper.

\section{RELATED WORK}
\label{sec2}
\subsection{Depth Improvement for Active Stereo Cameras}
The AS-based camera is widely used for indoor robotic applications due to its high accuracy and efficiency. However, it produces depth images with missing data when surfaces are highly reflective. To improve depth data quality, the HDR technique was developed \cite{zhang2009high, ekstrand2011autoexposure, yu2019adaptive, lin2017review, feng2018high}. It captures the pattern image with multiple camera exposures for stereo matching. However, the HDR method requires capturing images with up to 30 exposures, making it time-consuming. Moreover, the HDR techniques cannot handle the low SNR, where almost no light is reflected back to the camera, regardless of exposure length. Recently, with the great success of deep learning, some learning-based approaches have been developed to improve the depth data quality for AS-based cameras. These methods leveraged convolutional neural networks (CNN) to either enhance the single exposure-captured image \cite{liu2020optical}, or directly fill the missing areas in depth maps \cite{zhang2018deep, chai2020deep, sajjan2020clear, luo2021self}. However, single-view network-based methods learn a dataset prior that does not generalize well to other objects and reflective characteristics, and is not directly able to address low SNR from one viewpoint. In comparison, multi-view fusion \cite{wang2019highly, yang2021probabilistic} can provide higher levels of depth completion for reflective objects and varied scenes. The selection of viewpoints in multi-view fusion remains a critical step for practical use in the real world.

\subsection{Next-Best-View Prediction}
Active vision \cite{aloimonos1988active, chen2011active, bajcsy2018revisiting, zeng2020view}, and more specifically Next-Best-View (NBV) prediction, refers to camera viewpoint manipulation in order to collect useful information for various tasks at the next frame~\cite{isler2016information, daudelin2017adaptable, sanket2018gapflyt, wu2019plant, wu2021object, monica2021probabilistic, rebello2017autonomous, sock2020active, kiciroglu2020activemocap}. Among these works, the closest example to our application is 3D reconstruction \cite{isler2016information, daudelin2017adaptable, wu2019plant, wu2021object, monica2021probabilistic}. In \cite{isler2016information}, the authors proposed several formulations to quantify the information gain for the volumetric reconstruction of an object. The next best viewpoint is then selected by optimizing these formulations to discover new parts of the target object. These formulations were also employed in later works \cite{daudelin2017adaptable, wu2019plant, wu2021object, monica2021probabilistic} for object reconstruction or mapping of an environment. In \cite{wu2019plant} and \cite{monica2021probabilistic}, the authors leverage deep learning to predict the occupancy probabilities for unknown pixels or voxels, which were then used to guide NBV planning. However, all these works assumed complete depth acquisition from each camera viewpoint; missing depth caused by reflective surface and others were not considered.

\subsection{6D Pose Estimation for Reflective Objects}
As one of the most critical problems in robotics, 6D object pose estimation has been frequently addressed in the literature \cite{drost2010model, hinterstoisser2012model, rodrigues20126d, sundermeyer2018implicit, xiang2018posecnn, wang2019densefusion, chen2019pose, kozak2021data}. Among them, \cite{rodrigues20126d, chen2019pose, kozak2021data} tackled the object pose estimation for texture-less shiny parts using RGB images. However, as illustrated in \cite{yang2021robi, xiang2018posecnn, sundermeyer2018implicit}, RGB-only based methods can provide good 2D detection results but have low accuracy in terms of the final 6D pose. When high-quality depth data is provided, the object pose estimation performance can be significantly improved using refinement methods, such as iterative closest point (ICP). As a result, it is important to acquire good depth data in order to estimate the 6D poses for reflective objects.

\begin{figure*}[t]
\centering
  \includegraphics[width=0.92\linewidth]{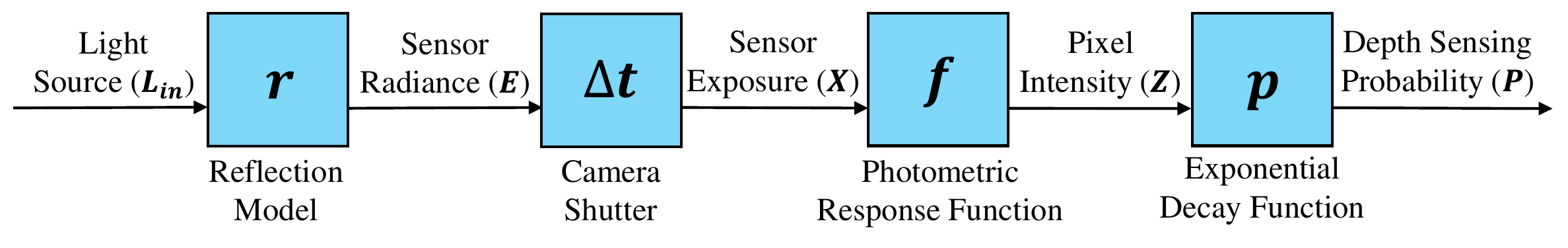}
\vspace{-0.4\baselineskip}  
\caption{Depth acquisition process for reflective surfaces with an AS-based camera.}
\label{fig3}
\vspace{-0.95\baselineskip}
\end{figure*}

\begin{figure}[t]
\centering
\begin{subfigure}{0.145\textwidth}
  \includegraphics[width=\linewidth]{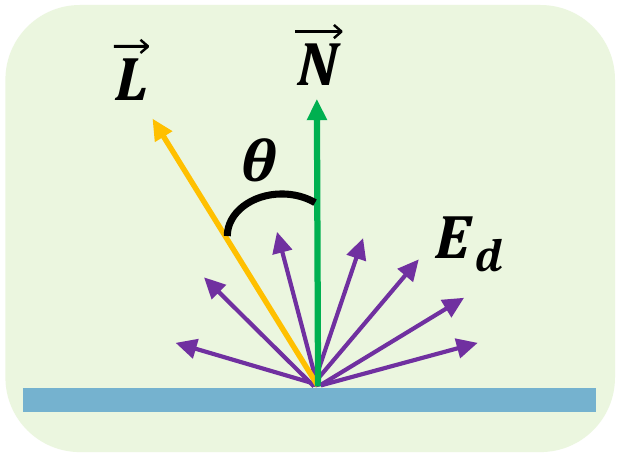}
  \vspace{-1.4\baselineskip}
  \caption{}
  \label{fig2a}
\end{subfigure}
\begin{subfigure}{0.145\textwidth}
    \includegraphics[width=\linewidth]{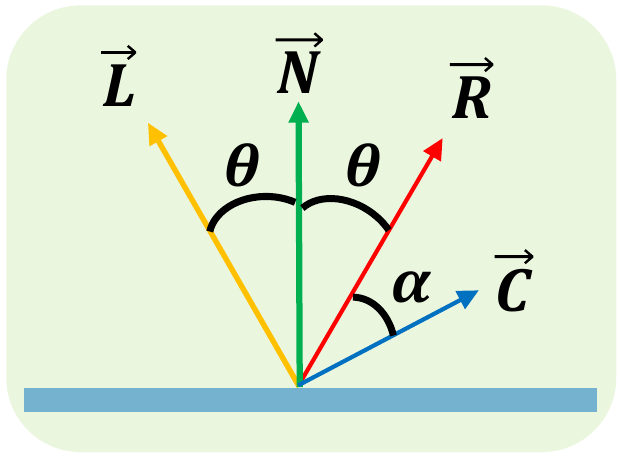}
    \vspace{-1.4\baselineskip}
    \caption{}
    \label{fig2b}
\end{subfigure}
\begin{subfigure}{0.15\textwidth}
    \includegraphics[width=\linewidth]{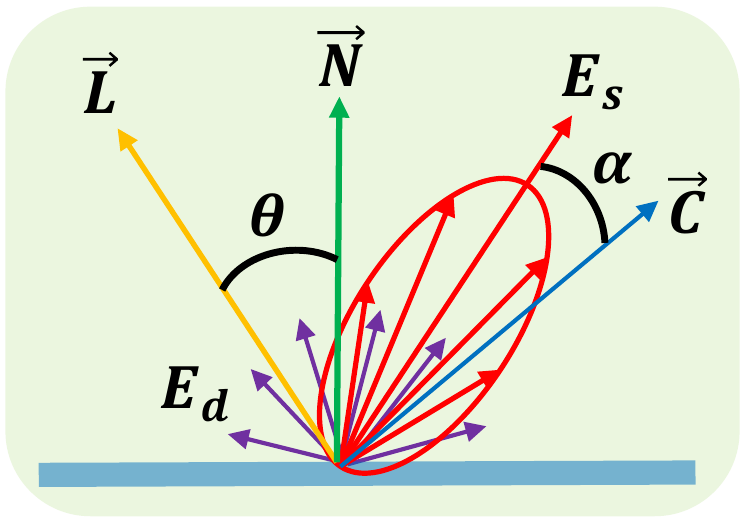}
    \vspace{-1.4\baselineskip}
    \caption{}
    \label{fig2c}
\end{subfigure}
\caption{Illustration of different cases for light reflection. (a) Diffuse reflection. (b) Specular reflection. (c) Phong reflection model.}
\label{fig2}
\vspace{-1.0\baselineskip}
\end{figure}

\section{Surface Reflection for Reflective Objects}
\label{sec3}
The depth acquisition of an AS-based camera is significantly influenced by the light sources, camera viewpoint, and surface characteristics (e.g., surface normal and materials). Figure \ref{fig3} illustrates the depth acquisition process for reflective surfaces with an AS-based camera. The radiance $\boldsymbol{E}$ represents the amount of light that comes into the camera lens per time unit, and the sensor exposure $\boldsymbol{X}$ is the total amount of energy that hits the pixel within the camera exposure time $\boldsymbol{\Delta t}$. The photometric response function then maps the exposure $\boldsymbol{X}$ to the pixel intensity $\boldsymbol{Z}$ in the pattern image. Finally, the depth-sensing probability $\boldsymbol{P}$ can be estimated based on the pixel intensity $\boldsymbol{Z}$.

Typically, for an AS-based camera, two light sources need to be considered: the ambient light with an intensity $\boldsymbol{L_{a}}$ and projector light with an intensity $\boldsymbol{L_{p}}$. Since the projector light is usually the dominating light source and the ambient light intensity, $\boldsymbol{L_{a}}$, is negligible in comparison, we define the total light intensity $\boldsymbol{L_{in}}$ as:
\begin{equation}
\label{equ3}
    \boldsymbol{L_{in}} = \boldsymbol{L_{a}} + \boldsymbol{L_{p}} \approx \boldsymbol{L_{p}}
\end{equation}

When the imaging object is a perfect Lambertian reflector, pure diffuse reflection happens. As demonstrated in Figure \ref{fig2a}, the surface reflects the incoming light with an intensity $\boldsymbol{L_{in}}$ by Lambert’s cosine law. The reflected light is determined only by the angle $\boldsymbol{\theta}$ between the direction of incident ray $\boldsymbol{\Vec{L}}$ and surface normal $\boldsymbol{\Vec{N}}$, and is independent on camera viewpoint:
\begin{equation}
\label{equ4}
    \boldsymbol{E_d} = \boldsymbol{L_{in}} \left( \boldsymbol{\Vec{L}} \cdot \boldsymbol{\Vec{N}} \right) = \boldsymbol{L_{in}} \cos\boldsymbol{\theta}, \;\;\; 0\leq\boldsymbol{\theta}\leq \pi/2
\end{equation}
where $\boldsymbol{E_d}$ is the diffuse radiance received by the camera. In comparison, when the light hits an ideal specular surface, the reflected ray $\boldsymbol{\Vec{R}}$ reflects off the mirror and propagates on the other side of the surface normal $\boldsymbol{\Vec{N}}$:
\begin{equation}
\label{equ41}
    \boldsymbol{\Vec{R}} = 2\boldsymbol{\left(\Vec{L} \cdot \Vec{N}\right)\Vec{N}} - \boldsymbol{\Vec{L}}
\end{equation}
The angle of the reflected ray $\boldsymbol{\Vec{R}}$ equals to the angle of incidence $\boldsymbol{\theta}$ (shown in Figure \ref{fig2b}). However, for a general reflective surface in the real world, specular reflections are distributed near the direction of the reflection ray $\boldsymbol{\Vec{R}}$. The reflected light received by a camera is determined by the light source, camera viewpoint, and surface glossiness. According to \cite{phong1975illumination}, the specular radiance $\boldsymbol{E_s}$ for a smooth surface is generally simulated as: 
\begin{equation}
\label{equ5}
    \boldsymbol{E_s} =
    \begin{cases}
    \boldsymbol{L_{in}} {\left(\boldsymbol{\Vec{R}} \cdot \boldsymbol{\Vec{C}}\right)}^{\boldsymbol{n}} =\boldsymbol{L_{in}} {\left(\cos\boldsymbol{\alpha}\right)}^{\boldsymbol{n}}, & 0\leq\boldsymbol{\alpha}\leq \pi/2\\
    0, & \pi/2 < \boldsymbol{\alpha}\leq \pi
    \end{cases} 
\end{equation}
where $\boldsymbol{n}$ is the glossiness parameter for the object material, which is larger for surfaces that are more mirror-like, and $\boldsymbol{\alpha}$ is the angle between the camera ray $\boldsymbol{\Vec{C}}$ and the perfectly reflected ray $\boldsymbol{\Vec{R}}$. When the angle $\boldsymbol{\alpha}$ is smaller, a stronger reflected signal, $\boldsymbol{E_s}$, is received by the camera. On the contrary, when $\boldsymbol{\alpha}$ is larger than $\pi/2$, the received light, $\boldsymbol{E_s}$, quickly drops to zero \cite{tan2017specular}.

In practice, there is no perfectly diffuse or specular material, and both diffuse and specular reflections exist simultaneously for all natural surfaces. In this work, we use the Phong reflection model  \cite{phong1975illumination}, as shown in Figure \ref{fig2c}, to represent the natural reflection properties for surfaces:
\begin{equation}
\label{equ6}
    \boldsymbol{E} = \boldsymbol{k_d E_d} + \boldsymbol{k_s E_s}
\end{equation}
where $\boldsymbol{E}$ represents the total radiance that the camera receives, $\boldsymbol{k_d}$ and $\boldsymbol{k_s}$ are diffuse and specular parameters, respectively. Larger $\boldsymbol{k_d}$ and smaller $\boldsymbol{k_s}$ indicate that the reflection is more close to diffuse reflection, and vice versa. To estimate the Phong parameters, we utilize a two-stage least square method. The details are described in Section \ref{sec32}.

\begin{figure}[t]
\centering
\begin{subfigure}{0.24\textwidth}
  \includegraphics[width=\linewidth]{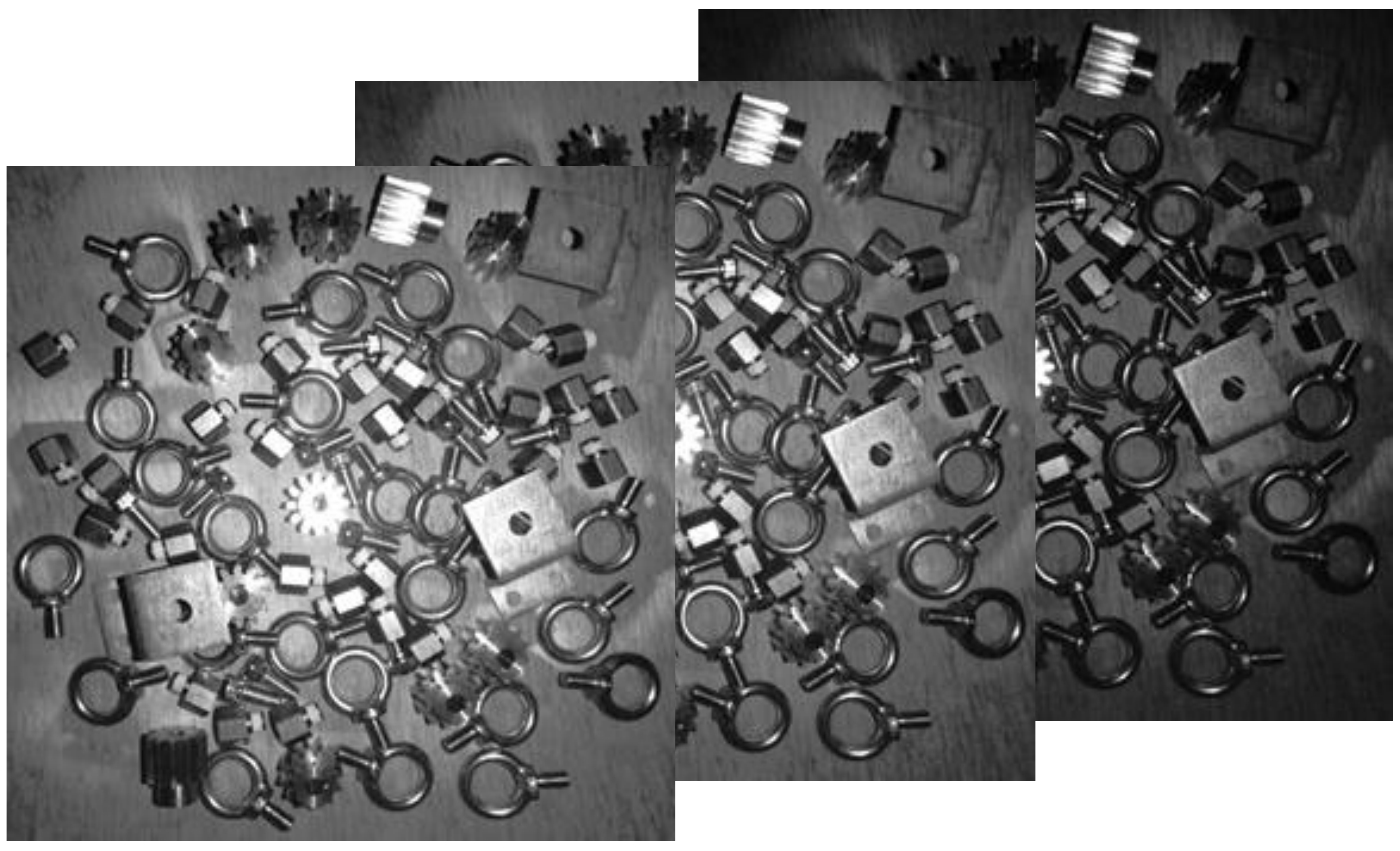}
  \vspace{-1.4\baselineskip}
  \caption{}
  \label{fig4a}
\end{subfigure}
\begin{subfigure}{0.20\textwidth}
    \includegraphics[width=\linewidth]{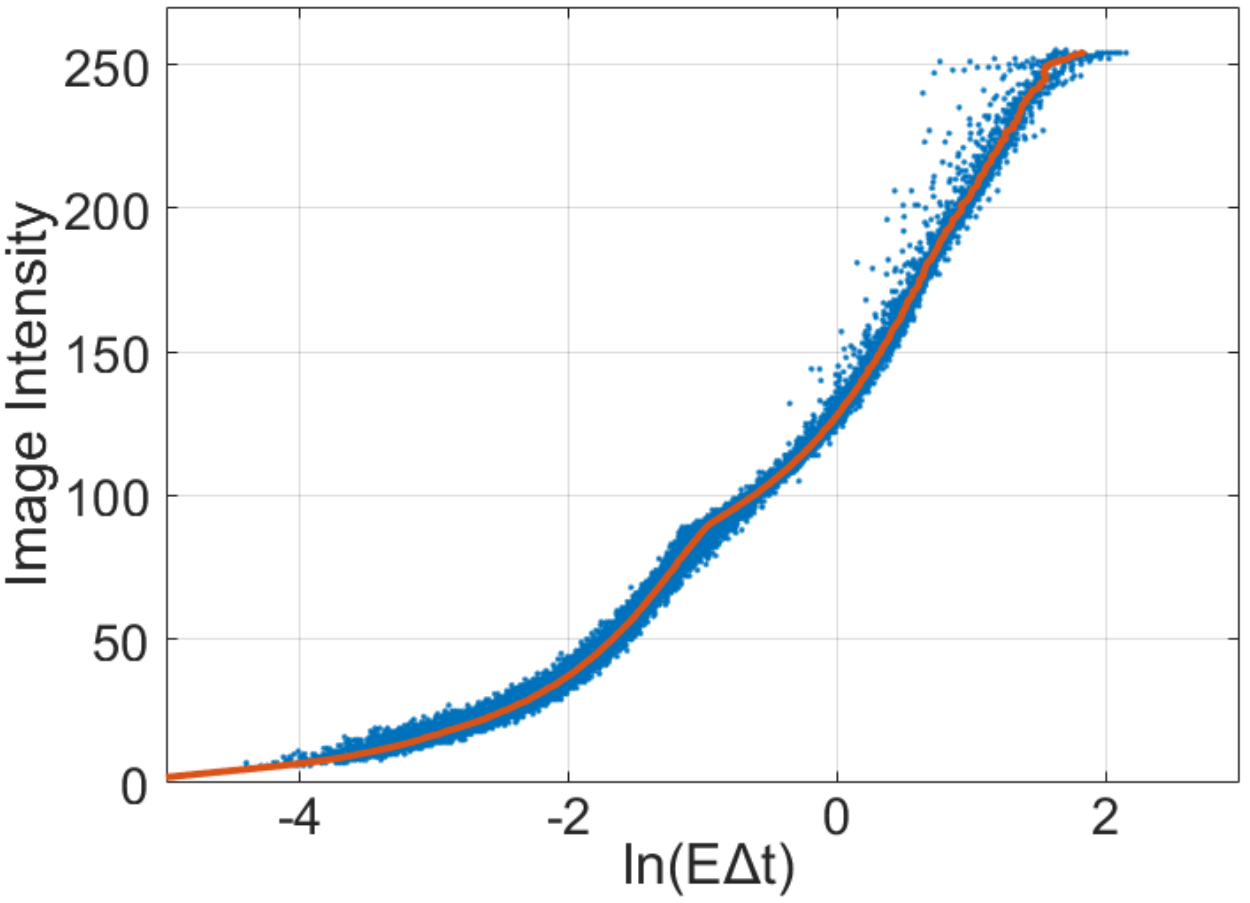}
    \vspace{-1.4\baselineskip}
    \caption{}
    \label{fig4b}
\end{subfigure}
\caption{Photometric response function estimation. (a) Captured input images. (b) The recovered function.}
\label{fig4}
\vspace{-0.8\baselineskip}
\end{figure}

With the identified Phong parameters, we can compute the expected image intensity $\boldsymbol{Z}$ using a photometric response function. As illustrated in Figure \ref{fig3}, after receiving the radiance $\boldsymbol{E}$, the camera captures an exposure $\boldsymbol{X}$ (product of radiance and exposure time, $\boldsymbol{E \Delta t}$). The photometric response function (illustrated in Figure \ref{fig4b}) then maps the exposure $\boldsymbol{X}$ to a digital number $\boldsymbol{Z}$, which is the intensity in the pattern image:
\begin{equation}
\label{equ_photometric}
    \boldsymbol{Z} = f(\boldsymbol{X})= f(\boldsymbol{E} \boldsymbol{\Delta t})
\end{equation}

And to sense the optimal depth data, it is crucial to avoid both image saturation and low SNR. Therefore, the pixel intensities for the object's surface are expected to come as close to saturation as possible without actually becoming saturated. For each pixel, the depth-sensing probability $\boldsymbol{P}$ increases with its intensity $\boldsymbol{Z}$, and quickly drops to zero if it is larger than $\boldsymbol{Z_{max}}$ (e.g., $255$). Many functions can provide this attribute, for convenience, we choose an exponential decay function to represent the depth-sensing probability $\boldsymbol{P}$ if the pixel intensity is $\boldsymbol{Z}$:
\begin{equation}
\label{equ10}
    \boldsymbol{P} \approx
    \begin{cases}
    \exp{\left(\frac{\boldsymbol{Z}-\boldsymbol{Z_{max}}}{\boldsymbol{\sigma}}\right)}, & \boldsymbol{Z_{min}} \leq \boldsymbol{Z} \leq \boldsymbol{Z_{max}}\\
      0, & \boldsymbol{Z} < \boldsymbol{Z_{min}} \;\;or\;\; \boldsymbol{Z} > \boldsymbol{Z_{max}}
    \end{cases}    
\end{equation}
where $\boldsymbol{\sigma}$ is a variable set by the user (by default $\boldsymbol{\sigma} = 100$). For the SLC, the depth-sensing probability is simply the $\boldsymbol{P}$. And for the ASC, which searches for correspondences between two cameras, it cannot measure the depth to the surface when there is image saturation or low SNR in the left or right image. Hence, we need to compute $\boldsymbol{P_L}$ and $\boldsymbol{P_R}$ for both left and right images using Equation (\ref{equ10}), and the final depth-sensing probability is: $\boldsymbol{P} = \boldsymbol{\min} \left(\boldsymbol{P_L}, \boldsymbol{P_R} \right)$.

\subsection{Photometric Response Function}
\label{sec31}
To recover the photometric response function for a camera (shown in Equation (\ref{equ_photometric})), we use a method presented in \cite{debevec2008recovering, zhang2017active}, which assumes that the function $f$ is monotonically increasing and its inverse $f^{-1}$ is well defined. By taking the natural logarithm, the inverse photometric response function $g(\boldsymbol{Z})$ is then defined as:
\begin{equation}
\label{equ2}
    g(\boldsymbol{Z}) = \ln f^{-1} (\boldsymbol{Z}) = \ln \boldsymbol{E} + \ln \boldsymbol{\Delta t}
\end{equation}

The input to the algorithm is a number of images taken from a static scene with different known exposure times $\boldsymbol{\Delta t}$. The function $g(\boldsymbol{Z})$ can be computed, up to a factor of scale, by minimizing a quadratic objective function. We refer readers to \cite{debevec2008recovering} for more details about this process. An example of captured input images and the recovered photometric response function is shown in Figure \ref{fig4}.

\begin{figure}[t]
\centering
\begin{subfigure}{0.14\textwidth}
  \includegraphics[width=\linewidth]{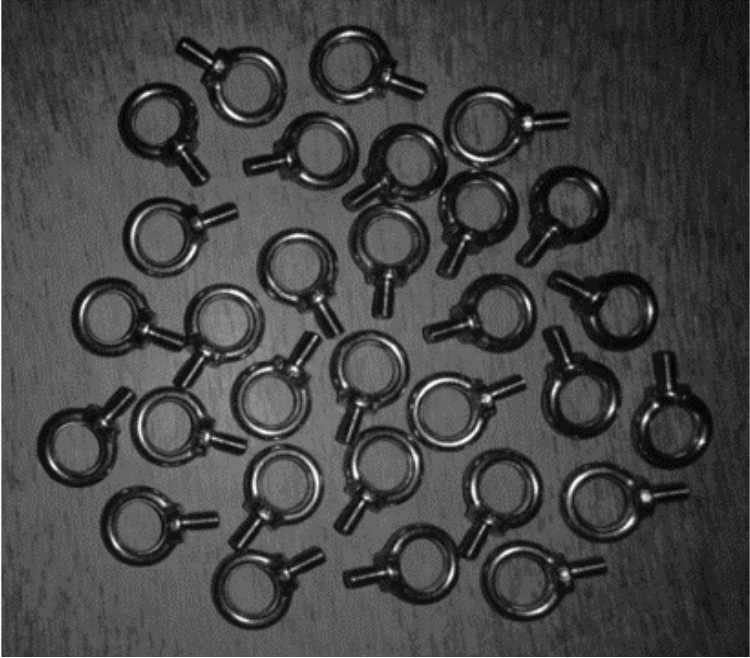}
  \vspace{-1.4\baselineskip}
  \caption{}
  \label{fig5a}
\end{subfigure}
\begin{subfigure}{0.165\textwidth}
    \includegraphics[width=\linewidth]{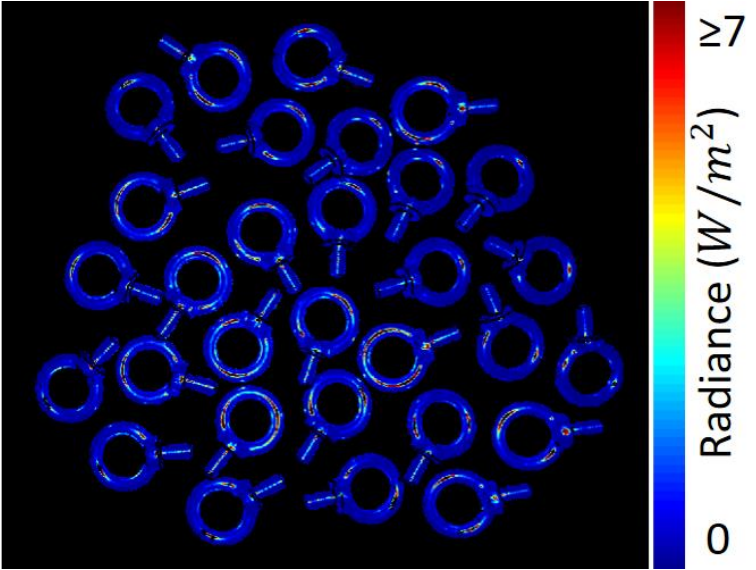}
    \vspace{-1.4\baselineskip}
    \caption{}
    \label{fig5b}
\end{subfigure}
\begin{subfigure}{0.14\textwidth}
    \includegraphics[width=\linewidth]{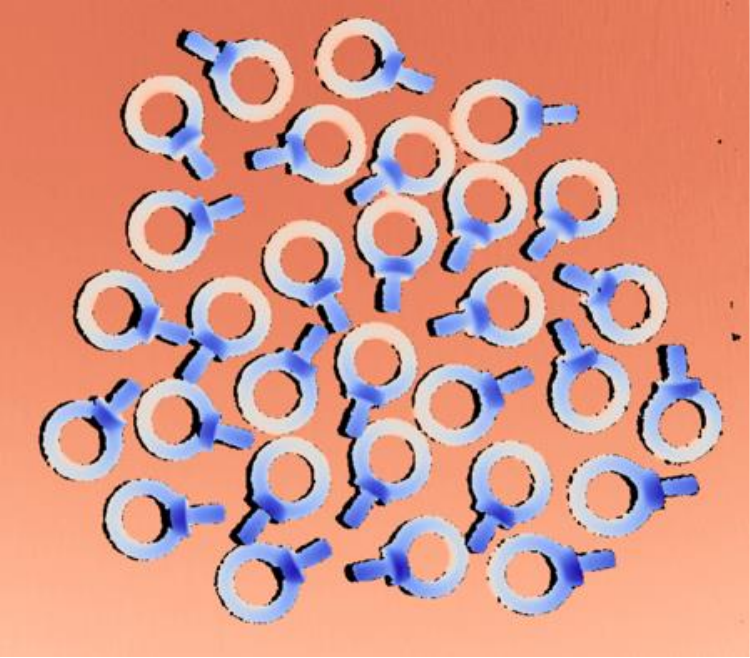}
    \vspace{-1.4\baselineskip}
    \caption{}
    \label{fig5c}
\end{subfigure}
\caption{Parameter Estimation for Phong Model. (a) Captured calibration image. (b) Recovered radiance map. (c) Captured depth map with the scanning spray.}
\label{fig6}
\end{figure}

\subsection{Parameter Estimation of Phong Model}
\label{sec32}
For each object material, we solve its Phong reflection parameters $\boldsymbol{\left[n, k_d, k_s\right]}$ with the known light intensity $\boldsymbol{L_{in}}$, incident ray $\boldsymbol{\Vec{L}}$, the pre-computed surface normal $\boldsymbol{\Vec{N}}$ and corresponding radiance $\boldsymbol{E}$. Specifically, we capture a static scene of the target object with two scans. In the first scan, we capture the image $\boldsymbol{I}_{calib}$ with the known exposure time $\boldsymbol{\Delta t}_{calib}$. A pure white pattern is projected on the scene for this capture, as shown in Figure \ref{fig5a}. We then compute the radiance $\boldsymbol{E}_{calib}$ for each pixel of the object reflective surface (Figure \ref{fig5b}) using the previously recovered photometric response function (Section \ref{sec31}):
\begin{equation}
\label{equ7}
    \boldsymbol{E}_{calib} = \exp \left(g\left(\boldsymbol{Z}_{calib}\right) - \ln \left( \boldsymbol{\Delta t}_{calib}\right)\right)
\end{equation}
where $\boldsymbol{Z}_{calib}$ is the pixel intensities of image $\boldsymbol{I}_{calib}$. For the second scan, we apply an anti-reflective scanning spray \cite{AESUB} on the parts to create the diffuse surfaces, so that the captured depth map $\boldsymbol{D}_{calib}$ (Figure \ref{fig5c}) can achieve its optimal accuracy with no missing depth data. The surface normal $\boldsymbol{\Vec{N}}_{calib}$ is then generated by applying Principal Component Analysis to the local neighborhood of each point in 3D space.

Considering that Equation (\ref{equ6}) is nonlinear, inspired by \cite{tan2017specular}, we solve for the diffuse parameter $\boldsymbol{k_d}$ and specular parameters $\boldsymbol{n}$, $\boldsymbol{k_s}$ separately. We first compute the diffuse parameter $\boldsymbol{k_d}$ via a least-squares solution on Equation (\ref{equ4}) with the sub-data that $\boldsymbol{\alpha}$ is larger than $\pi/2$ (only diffuse reflection exists, $\boldsymbol{E}_{calib} = \boldsymbol{E_d}$). Then, the sub-data with $\boldsymbol{\alpha} < \pi/2$ can be used to determine the specular parameters by subtracting estimated diffuse reflection:
\begin{equation}
\label{equ8}
    \boldsymbol{k_s} {(\cos\boldsymbol{\alpha})}^{\boldsymbol{n}} \boldsymbol{L_{in}}= \boldsymbol{Q}
\end{equation}
where $\boldsymbol{Q} = \boldsymbol{E}_{calib} - \boldsymbol{k_d} \boldsymbol{L_{in}} \cos\boldsymbol{\theta}$. We take the natural logarithm function on both sides of Equation (\ref{equ8}):
\begin{equation}
\label{equ9}
    \ln{\boldsymbol{k_s}} + \boldsymbol{n} \ln{(\cos\boldsymbol{\alpha})} = \ln{\frac{\boldsymbol{Q}}{\boldsymbol{L_{in}}}}
\end{equation}
The parameters $\boldsymbol{\left[n, k_s\right]}$ can be finally estimated using another least-squares solution. Figure \ref{fig6} shows our estimated Phong parameters for different surface materials.

\begin{figure}[t]
\centering
\begin{subfigure}{0.15\textwidth}
  \includegraphics[width=\linewidth]{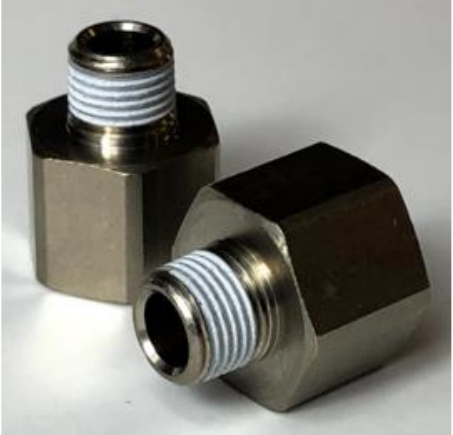}
  \vspace{-1.4\baselineskip}
  \captionsetup{justification=centering}
  \caption{$\boldsymbol{k_d}=0.037, \boldsymbol{k_s}=0.74, \boldsymbol{n}=19.9$}
  \label{fig6a}
\end{subfigure}
\begin{subfigure}{0.15\textwidth}
    \includegraphics[width=\linewidth]{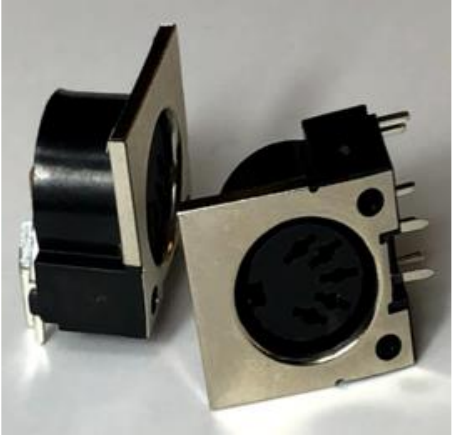}
    \vspace{-1.4\baselineskip}
    \captionsetup{justification=centering}
    \caption{$\boldsymbol{k_d}=0.04, \boldsymbol{k_s}=0.82, \boldsymbol{n}=38.9$}    
    \label{fig6b}
\end{subfigure}
\begin{subfigure}{0.15\textwidth}
    \includegraphics[width=\linewidth]{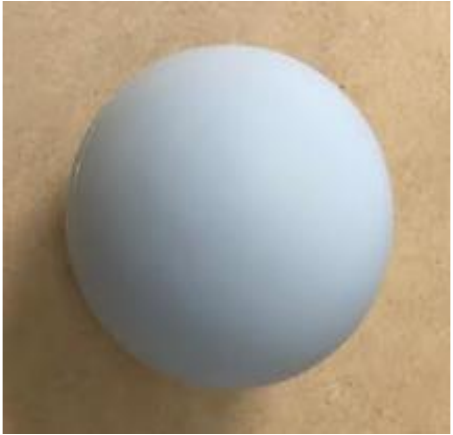}
    \vspace{-1.4\baselineskip}
    \captionsetup{justification=centering}
    \caption{$\boldsymbol{k_d}=0.45, \boldsymbol{k_s}=0.02, \boldsymbol{n}=8.65$} 
    \label{fig6c}
\end{subfigure}
\caption{Phong parameters for different object materials. (a)\&(b) Reflective objects "Tube Fitting" and "DIN Connector" (Figures are from \cite{yang2021probabilistic}). For an object with hybrid materials, we estimated the parameters for metallic surfaces only. (c) The object with a matte surface.}
\label{fig6}
\end{figure}

\section{Next-Best-View Prediction}
\label{sec4}
The next-best-view planner selects the camera viewpoint $\boldsymbol{v}^*$ from a set of candidate viewpoints $\boldsymbol{\{V\}}$ by maximizing the information gain. In our work, information gain is the amount of information that a viewpoint can provide for completing the reference view's missing depth data on reflective surfaces. In Section \ref{sec3}, we propose the surface reflection model for this computation, and the information gain $\boldsymbol{G_i}$ for a candidate viewpoint $\boldsymbol{v_i}$ can be defined as:
\begin{equation}
\label{equ12}
    \boldsymbol{G_i} = \sum_{\boldsymbol{u}} h\left(\boldsymbol{u}, \boldsymbol{\check{D}}, \boldsymbol{\check{\Vec{N}}}, \boldsymbol{v_i}\right)
\end{equation}
where the function $h\left(\boldsymbol{u}, \boldsymbol{\check{D}}, \boldsymbol{\check{\Vec{N}}}, \boldsymbol{v_i}\right)$ is our reflection model for predicting the depth-sensing probability using equations (\ref{equ3})-(\ref{equ10}). $\boldsymbol{\check{D}}$ and $\boldsymbol{\check{\Vec{N}}}$ are the hypotheses of depth and surface normal on the missing depth pixels ${\boldsymbol{u}}$, which can be considered as the prior information of the environment. In \cite{wu2019plant, monica2021probabilistic}, authors used deep learning-based methods to learn these priors for NBV prediction guidance. However, learning-based approaches require a large amount of training data and are prone to over-fitting to a particular dataset. Moreover, these methods took the entire environment as the exploration target, which may misguide the camera into the non-object area. In this work, we take advantage of the 3D object CAD model by first employing an RGB-based pose estimator to estimate initial object pose hypotheses $\{\boldsymbol{q}\}$. The hypotheses of depth $\{\boldsymbol{\check{D}}\}$ and surface normal $\{\boldsymbol{\check{\Vec{N}}}\}$ can be then inferred from the 3D object CAD model.

\begin{figure}[t]
\centering
\begin{subfigure}{0.155\textwidth}
  \includegraphics[width=\linewidth]{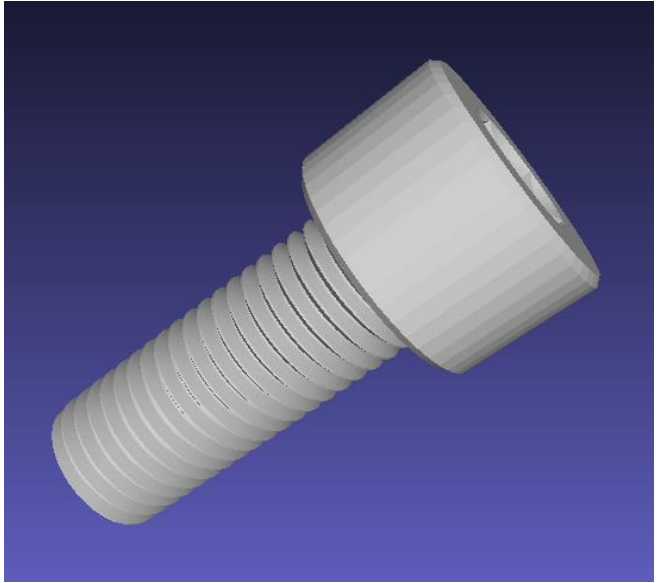}
  \vspace{-1.4\baselineskip}
  \caption{}
  \label{fig7a}
\end{subfigure}
\begin{subfigure}{0.155\textwidth}
    \includegraphics[width=\linewidth]{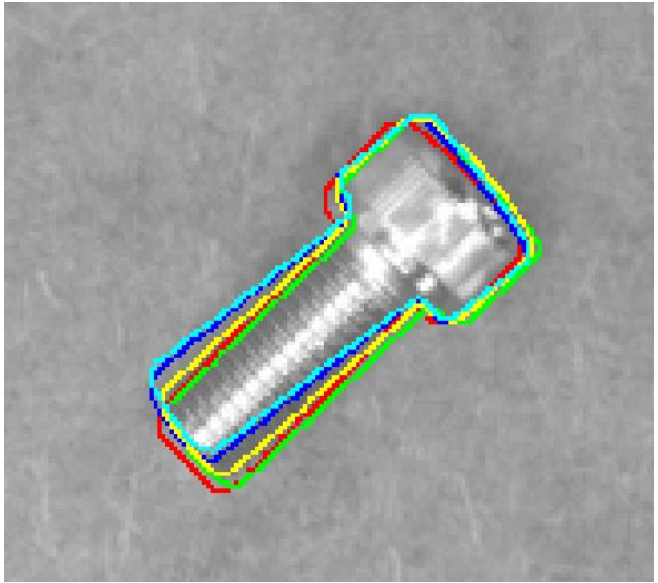}
    \vspace{-1.4\baselineskip}
    \caption{}
    \label{fig7b}
\end{subfigure}
\begin{subfigure}{0.155\textwidth}
    \includegraphics[width=\linewidth]{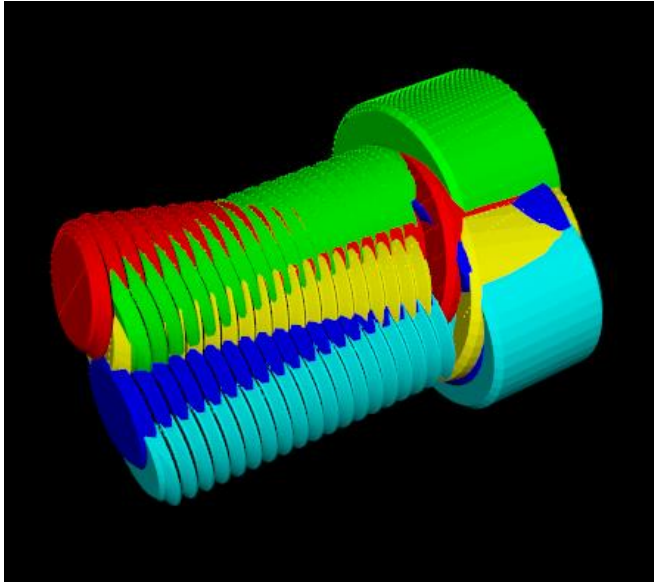}
    \vspace{-1.4\baselineskip}
    \caption{}
    \label{fig7c}
\end{subfigure}
\caption{The Line2D pose estimator for the object "Chrome Screw". (a) Object CAD model. (b)\&(c) Pose estimation results, illustrated in 2D and 3D space, respectively.}
\label{fig7}
\end{figure}

\begin{algorithm}[t]
   \SetKwInOut{Input}{Input}
  \SetKwInOut{Output}{Output}
  \DontPrintSemicolon
  % Option
  \Input{Grayscale image $\boldsymbol{I_o}$; depth map $\boldsymbol{D_o}$; a set of candidate viewpoints $\boldsymbol{\{V\}}$.}
  \Output{Completed depth map $\boldsymbol{D_o^{'}}$.}
 Line2D pose estimation:  $\boldsymbol{I_o}\boldsymbol{\xrightarrow{}\{q\}}$;\;
 Create prior information: $\boldsymbol{\{q\}}\xrightarrow{}\boldsymbol{\{\check{D}, \check{\Vec{N}}\}}$;\;
 Compute $\boldsymbol{G_i}$ for each $\boldsymbol{v_i}$ in $\boldsymbol{\{V\}}$ using equation (\ref{equ13});\;
 Determine the NBV $\boldsymbol{v^*}$, move the camera;\;
 Capture $\boldsymbol{D^*}$ from $\boldsymbol{v^*}$, apply fusion:
$\boldsymbol{D^*}, \boldsymbol{D_o} \xrightarrow{} \boldsymbol{D_o^{'}}$;\;
 Check if the termination condition is fulfilled;\;
 If condition is not fulfilled: Repeat steps from 3-6.
 \caption{Next-Best-View Prediction for Depth Completion on Reflective Objects}
 \label{alg1}
\end{algorithm}

In particular, we adopt the template matching-based approach, Line2D \cite{hinterstoisser2012model}. This method generates multi-view multi-scale object templates from the CAD model, and exploits the gradient response on RGB or grayscale images for detection in run-time. To better serve the template matching algorithm, we capture an additional image $\boldsymbol{I_o}$ with the camera projector off, and use a high exposure time to obtain optimal contrast for objects. Figure \ref{fig7} shows an example of the captured image and Line2D result on reflective objects. Due to the ambiguity at the 2D space, estimated 6D poses generally have large uncertainties, and each object may be matched with multiple templates. In other words, each pose cluster in 2D encapsulates the current belief about the object pose, which can be used to evaluate its expected information gain for a new viewpoint $\boldsymbol{v_i}$. Hence, we define the information gain by considering all pose hypotheses in the cluster:
\begin{align}
\label{equ13}
    \boldsymbol{G_i} &= \sum_{k=0}^{K-1} P\left(\boldsymbol{q}_k\right)\left(\sum_{\boldsymbol{u}} h\left(\boldsymbol{u}, \boldsymbol{\check{D}}_k, \boldsymbol{\check{\Vec{N}}}_k, \boldsymbol{v_i}\right) \right) \\
    &= \sum_{k=0}^{K-1} \frac{\exp{\left(\boldsymbol{c}_k\right)}}{\sum \exp{\left(\boldsymbol{c}_k\right)}}\left(\sum_{\boldsymbol{u}} h\left(\boldsymbol{u}, \boldsymbol{\check{D}}_k, \boldsymbol{\check{\Vec{N}}}_k, \boldsymbol{v_i}\right) \right)
\end{align}
where $\boldsymbol{\check{D}}_k$ and $\boldsymbol{\check{\Vec{N}}}_k$ are inferred from the object pose $\boldsymbol{q}_k$ and 3D CAD model. $P\left(\boldsymbol{q}_k\right)$ is the discrete probability of a pose hypothesis, which can be approximated using the softmax function with the Line2D confidence score $\boldsymbol{c}_k$. We compute the information gain for each candidate viewpoint, and the next-best-view $\boldsymbol{v}^*$ can be determined:
\begin{equation}
\label{equ14}
    \boldsymbol{v}^* = \arg\max \boldsymbol{G_i}
\end{equation}
The depth completion can be halted based on user needs, i.e., after a fixed number of iterations or when the highest expected information gain of a subsequent view falls below a user-defined threshold $\boldsymbol{G_{\tau}}$:
\begin{equation}
\label{equ15}
    \boldsymbol{G_i} < \boldsymbol{G_{\tau}}
\end{equation}
The architecture of NBV for depth completion on reflective objects is shown in Algorithm \ref{alg1}. It begins with the creation of the prior information, $\boldsymbol{\check{D}}, \boldsymbol{\check{\Vec{N}}}$, using the Line2D pose estimator and the grayscale image $\boldsymbol{I_o}$, from the reference viewpoint. For each iteration, the robot moves the camera to the predicted NBV $\boldsymbol{v}^*$ and captures a new depth map $\boldsymbol{D^*}$. To complete the reference view's depth map $\boldsymbol{D_o}$, we employ a volumetric fusion method \cite{yang2021probabilistic} and project the fused 3D data to the reference viewpoint at each iteration.

\begin{table*}[t]
\resizebox{\textwidth}{!}{
\begin{tabular}{|c|c|c|c|c|c|c|c|}
\hline
\multirow{2}{*}{Method} & \multicolumn{7}{c|}{Depth Completion Percentage (\%)}                                      \\ \cline{2-8} 
                        & Tube Fitting  & Chrome Screw  & Eye Bolt  & Gear & Zigzag & Din Connector & Dsub Connector \\ \hline
Random                  & 41.3          & 31.5          &      46.1     &   47.6   &    \textbf{65.2}    &     40.7          &       \textbf{48.0}         \\ \hline
Maximum Distance        & 47.5          & 28.9          &     48.2      &  41.0    &    64.2    &       34.2        &       35.9         \\ \hline
Proposed NBV            & \textbf{61.2} & \textbf{35.5} & \textbf{50.2} &  \textbf{54.2}    &    61.8    &      \textbf{49.9}         &     41.6           \\ \hline
\end{tabular}}
\caption{Depth completion results for reference viewpoint's depth map, evaluated with the metric of depth completion percentage. The maximum number of viewpoints is set to 3.}
 \vspace{-0.8\baselineskip}
\label{tab1}
\end{table*}

\begin{table*}[t]
\resizebox{\textwidth}{!}{
\begin{tabular}{|c|c|c|c|c|c|c|c|}
\hline
\multirow{2}{*}{Method} & \multicolumn{7}{c|}{Correct Detection Rate in \%, (ADD in mm)}                                      \\ \cline{2-8} 
                        & Tube Fitting & Chrome Screw & Eye Bolt     & Gear         & Zigzag       & Din Connector & Dsub Connector \\ \hline
Single View             & 72.4, (1.73) & 63.8, (1.42) & 72.6, (1.37) & 68.3, (1.83) & 45.3, (1.22) & 16.1, (1.47)  & 39.6, (1.74)   \\ \hline
Random                  & 75.5, (1.35) & 72.4, (1.19) & 72.6, (1.14) & 77.5, (1.43) & 45.3, (0.89) & \textbf{18.9}, (1.22)  & 46.1, (\textbf{1.39})   \\ \hline
Maximum Distance        & 75.7, (1.19) & 67.3, (\textbf{1.16}) & \textbf{73.5}, (1.22) & 75.2. (1.61) & 45.3, (\textbf{0.88}) & 17.6, (1.25)  & 45.2, (1.40)   \\ \hline
Proposed NBV            & \textbf{76.2}, (\textbf{1.04}) & \textbf{76.7}, (1.20) & 72.1, (\textbf{1.03}) & \textbf{78.4}, (\textbf{1.38}) & 45.3, (0.91) & \textbf{18.9}, (\textbf{1.20})  & \textbf{47.5}, (1.45)   \\ \hline
\end{tabular}}
\caption{Object pose estimation results for reference viewpoint's depth map, evaluated with the metrics of correct detection rate and ADD error. The maximum number of viewpoints is set to 3.}
 \vspace{-0.8\baselineskip}
\label{tab2}
\end{table*}

\section{EXPERIMENTS}
\label{sec5}
\subsection{Datasets and Experimental Setup}
\begin{figure}[t]
\centering
\begin{subfigure}{0.155\textwidth}
  \includegraphics[width=\linewidth]{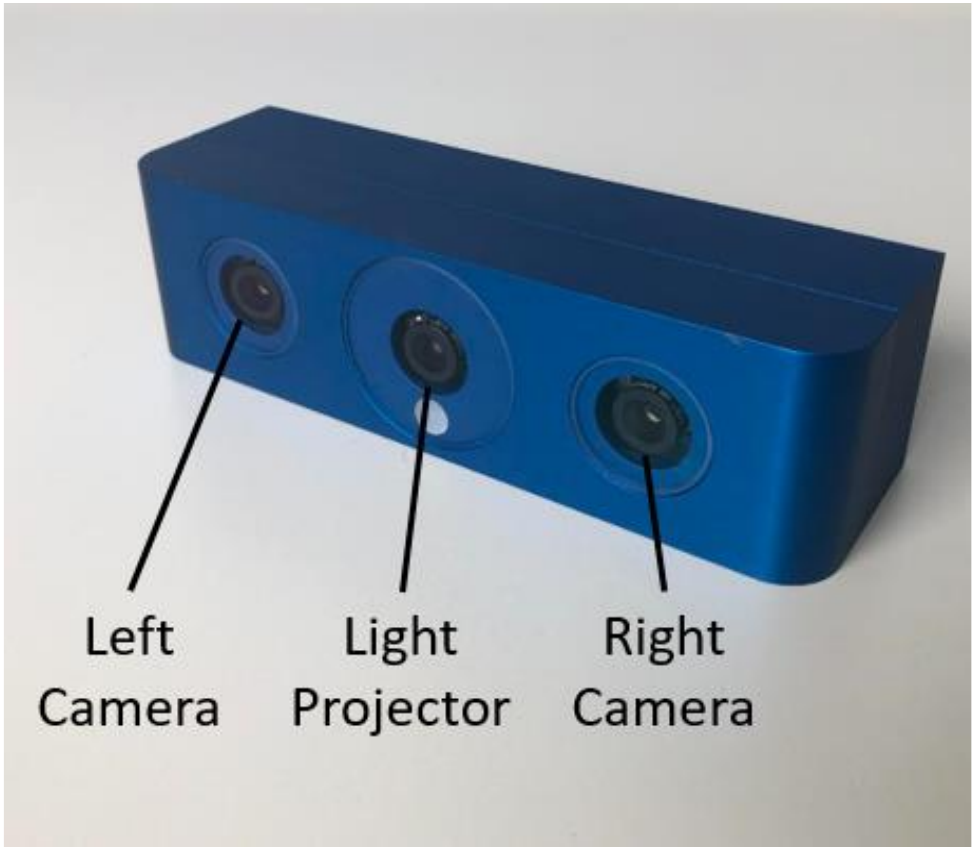}
  \vspace{-1.4\baselineskip}
  \caption{}
  \label{fig8a}
\end{subfigure}
\begin{subfigure}{0.155\textwidth}
    \includegraphics[width=\linewidth]{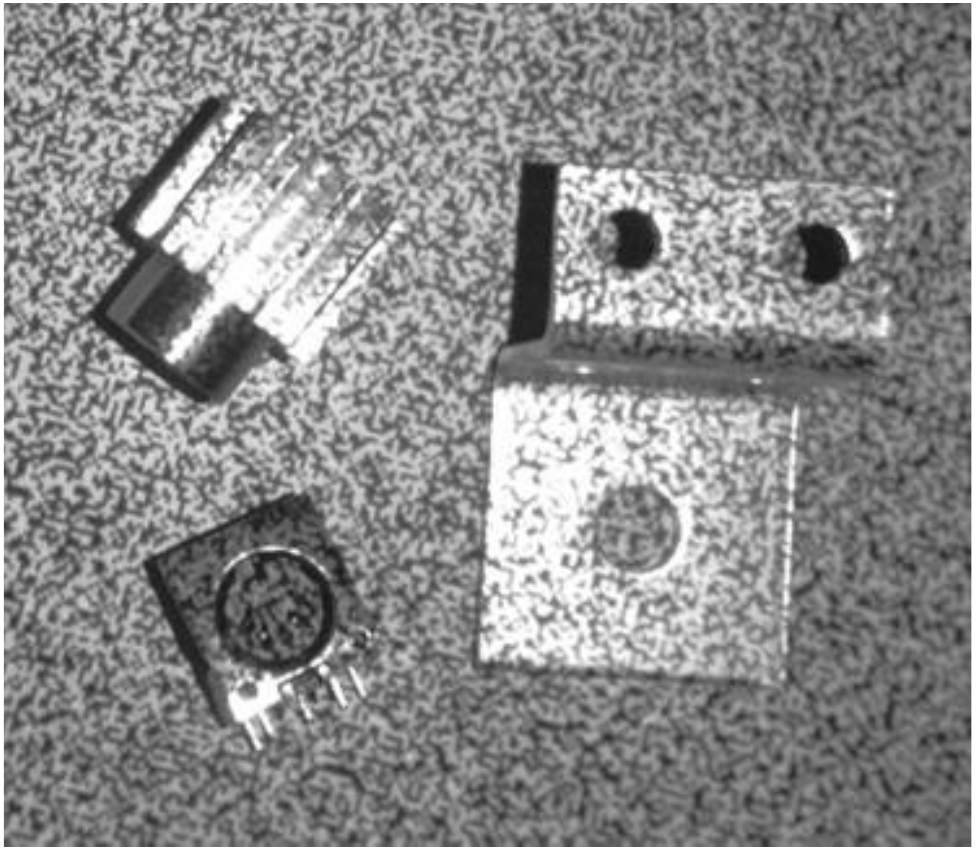}
    \vspace{-1.4\baselineskip}
    \caption{}
    \label{fig8b}
\end{subfigure}
\begin{subfigure}{0.155\textwidth}
    \includegraphics[width=\linewidth]{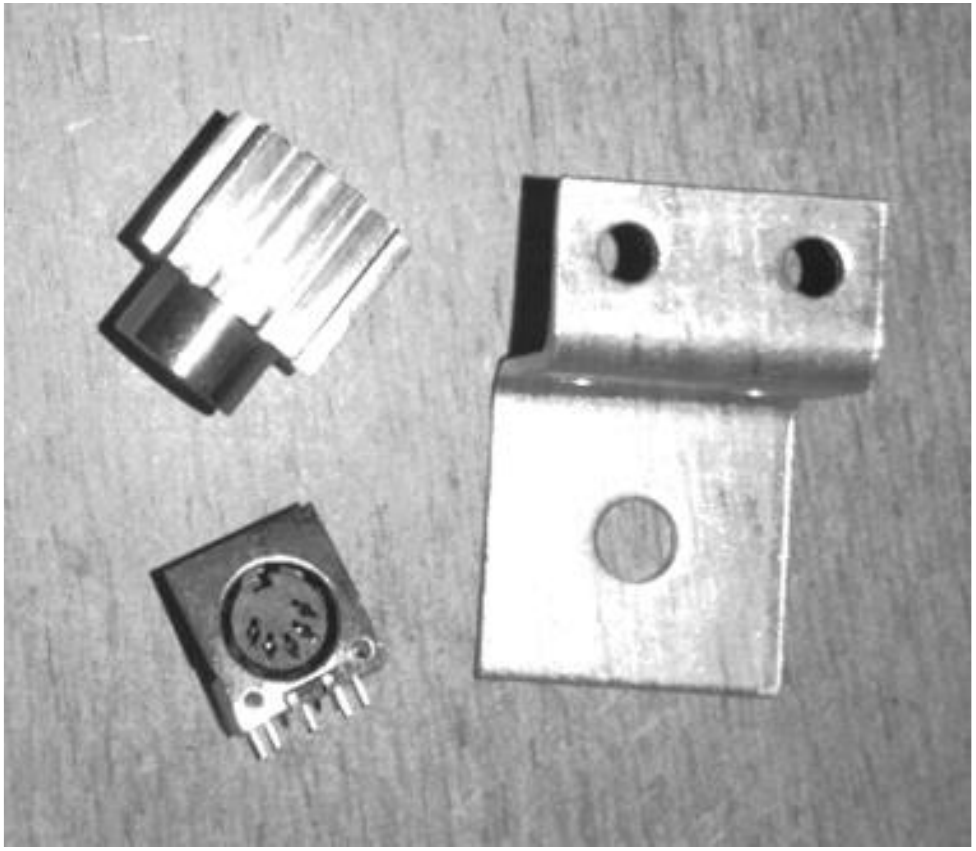}
    \vspace{-1.4\baselineskip}
    \caption{}
    \label{fig8c}
\end{subfigure}
\caption{(a) The Ensenso camera used for experiments. (b) Pattern image with random dots. (c) White pattern image.}
\label{fig8}
\end{figure}

In our experiments, we use a high-cost Ensenso ASC camera \cite{ENSENSO}, as shown in Figure \ref{fig8a}, and evaluate our NBV pipeline on ROBI dataset \cite{yang2021robi}, which was captured using the same sensor. The ROBI dataset contains seven highly reflective objects with a multi-view data acquisition setup. The biggest advantage of this dataset is that the ground truth depth maps are provided for evaluating the reconstruction and depth completion tasks.

To calibrate the camera's photometric response function (Section \ref{sec31}) and the Phong parameters (Section \ref{sec32}) of object's metallic surfaces, the pure white pattern projected images $\boldsymbol{I}_{calib}$ are required. We capture these images by employing the built-in "FlexView-16" functionality of Ensenso camera\footnote{For each acquisition, a total of 16 stereo pairs are captured sequentially, while the pseudorandom dots are shifted spatially in the projected pattern. We apply the maximum filter over the temporal domain of 16 shots for each pixel to get the white pattern image. More details of the "FlexView-16" technology can be found in \cite{ENSENSO}.}. Examples of random pattern and white pattern images are shown in Figure \ref{fig8b} and \ref{fig8c}, respectively.

\begin{figure}[t]
\centering
\begin{subfigure}{0.225\textwidth}
  \includegraphics[width=\linewidth]{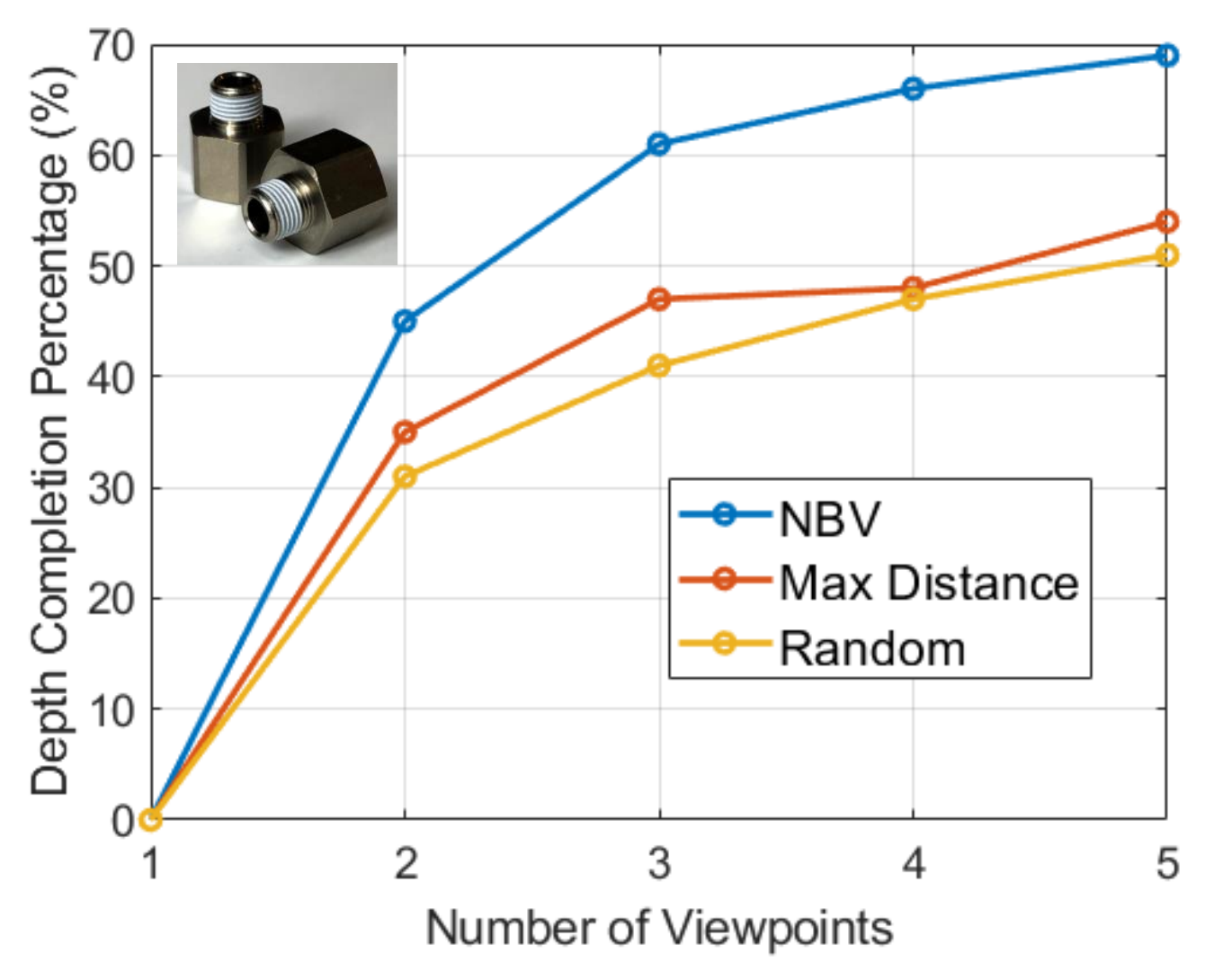}
  \vspace{-1.4\baselineskip}
  \caption{}
  \label{fig9a}
\end{subfigure}
\begin{subfigure}{0.225\textwidth}
    \includegraphics[width=\linewidth]{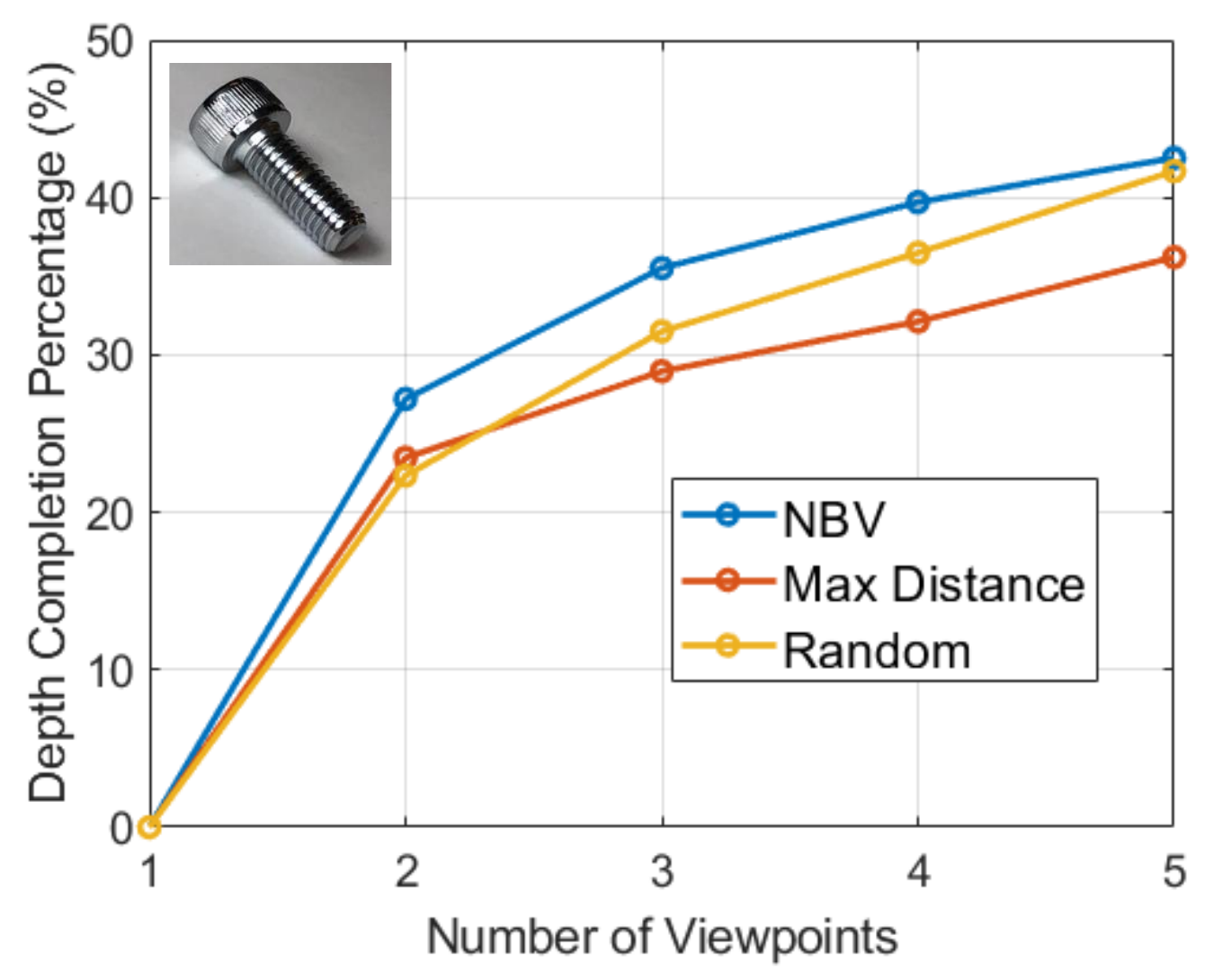}
    \vspace{-1.4\baselineskip}
    \caption{}
    \label{fig9b}
\end{subfigure}
\caption{Depth completion results of proposed NBV system compared to two different baselines: (a) Object "Tube Fitting". (b) Object "Chrome Screw".}
\label{fig9}
\vspace{-0.8\baselineskip}
\end{figure}

\subsection{Evaluation Metrics and Baselines}
Both depth completeness and depth-based pose estimation are evaluated on each Line2D detected object. We define depth completeness as the fraction of the number of recovered pixels over the total number of missing pixels in the reference view's depth map $\boldsymbol{D_o}$. A pixel is determined as recovered if its depth error is less than $2\;mm$ compared with the ground truth depth. To evaluate the object pose estimation, we take the Line2D pose estimation results as the initial object pose guess and apply ICP on multi-view fused depth maps for refinement. Pose accuracy for all objects is evaluated with the average distance (ADD) metric, proposed in \cite{hinterstoisser2012model} and used for evaluation in \cite{sundermeyer2018implicit, xiang2018posecnn, wang2019densefusion, hodan2018bop}. A refined pose is counted as correct if its ADD is below 10\% of the object diameter.

We test our NBV system against two baselines, which can be easily employed by a non-expert human operator. The first baseline, "Random", selects random viewpoints from the set of candidate viewpoints. The second baseline, "Maximum Distance", moves the camera to the viewpoint of the furthest distance from previous viewpoints.

\subsection{Results}
\begin{figure}[t]
\centering
  \includegraphics[width=0.95\linewidth]{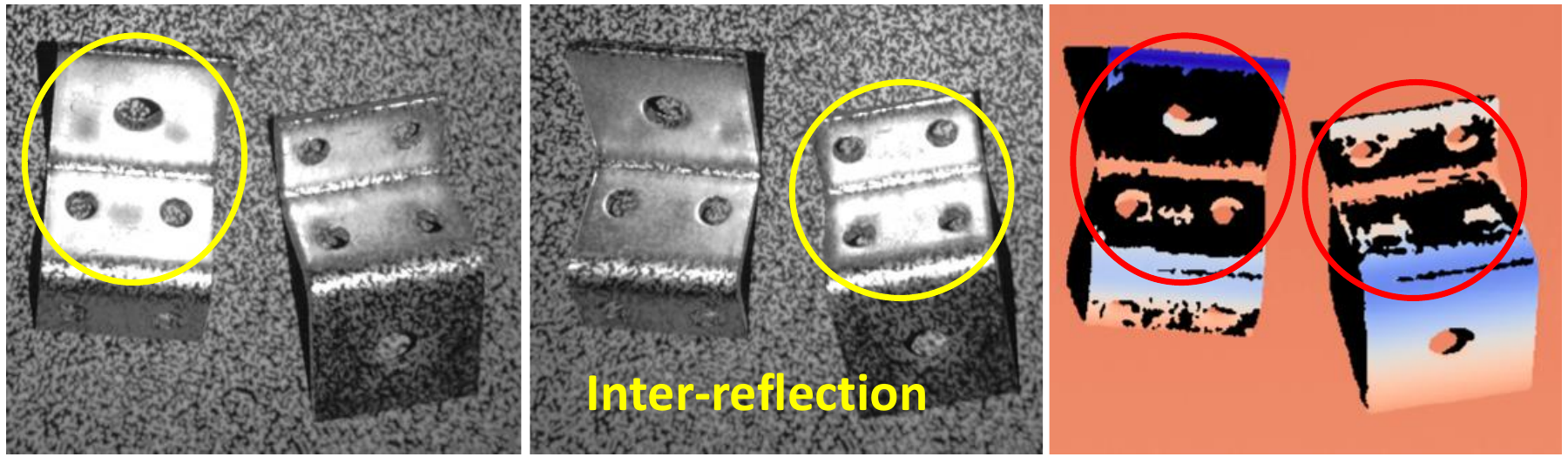}
\caption{Missing depth data on object "Zigzag", caused by inter-reflection.}
\vspace{-1.0\baselineskip}
\label{fig10}
\end{figure}

Table \ref{tab1} shows the depth completion results on different objects from the ROBI dataset. To obtain the results, the maximum number of viewpoints is set to 3. For objects with non-complex geometries, such as "Tube Fitting" (Figure \ref{fig6a}), our approach outperforms the other two baselines by a large margin. This is because the missing depth problem is mainly caused by image saturation or low SNR. Figure \ref{fig9} further demonstrates this behavior as the accumulation of viewpoints. Compared to the baselines, our proposed NBV approach requires fewer viewpoints to achieve the same level of depth completeness. However, our framework performs worse when objects have complex shapes, including large concavities (e.g., "Zigzag"). This is likely due to the inter-reflection (light is reflected within the object surfaces before returning to the camera, shown in Figure \ref{fig10}), which is not modeled in our work. To overcome this problem, ray tracing over the object surfaces should be considered. It is also noteworthy that when the object has only a small portion of reflective materials (e.g., "DIN Connector", shown in Figure \ref{fig6b}), our framework does not demonstrate a notable advantage. This is because the missing depth problem may be caused by other factors, such as black absorptive materials, which are notoriously difficult for active stereo illumination.

We present the object pose estimation results in table \ref{tab2}. It can be seen that, when more complete depth maps are provided, the pose errors (ADD) will be significantly reduced. This is particularly important for robotic applications that require highly accurate 6D object poses, such as robot grasping.

\section{CONCLUSION}
\label{sec6}
In this paper, we propose a next-best-view system for completing depth data on highly reflective objects. Based on the active stereo camera, we first explicitly model the specular reflection of reflective surfaces with the Phong reflection model and a photometric function. We then apply an RGB-based object pose estimator to provide a scene prior for predicting the next best viewpoint. We evaluate our method on a challenging dataset, and the performance of our approach outperforms two strong baselines when objects have non-complex shapes. In future work, we will investigate the use of ray tracing-based techniques for solving the inter-reflection problem.


%%%%%%%%%%%%%%%%%%%%%%%%%%%%%%%%%%%%%%%%%%%%%%%%%%%%%%%%%%%%%%%%%%%%%%%%%%%%%%%%

%%%%%%%%%%%%%%%%%%%%%%%%%%%%%%%%%%%%%%%%%%%%%%%%%%%%%%%%%%%%%%%%%%%%%%%%%%%%%%%%

%%%%%%%%%%%%%%%%%%%%%%%%%%%%%%%%%%%%%%%%%%%%%%%%%%%%%%%%%%%%%%%%%%%%%%%%%%%%%%%%

%%%%%%%%%%%%%%%%%%%%%%%%%%%%%%%%%%%%%%%%%%%%%%%%%%%%%%%%%%%%%%%%%%%%%%%%%%%%%%%%

\bibliographystyle{ieeetr}
\bibliography{root.bbl}

\end{document}